%% file: iclr2027_conference.tex
\documentclass{article} 
\usepackage{iclr2027_conference,times}

\input{math_commands.tex}

\usepackage{hyperref}
\usepackage{url}
\usepackage{graphicx}
\usepackage{booktabs}
\usepackage{tikz}
\usepackage{adjustbox}
\usepackage{float}

\title{Titan-Llama: Retrofitting Pre-trained LLMs with External Neural Memory Modules}

\author{Sarah Pan \\
Massachusetts Institute of Technology\\
\texttt{sarahpan@mit.edu} \\
\And
Ali Behrouz \\
Cornell University \\
\texttt{ab2947@cornell.edu} \\
}

\iclrfinalcopy 
\begin{document}

\maketitle

\begin{abstract}
Transformers are efficient to pre-train at scale because attention permits parallel computation across sequence positions, but long-context inference requires increasingly large KV caches.
Efficient recurrent alternatives reduce the asymptotic complexity of inference, but their sequential dependence limits token-wise parallelism during training, making them harder to scale.
Further, the current paradigm of extending long context tasks creates competing notions of retaining information precisely and keeping long-context inference efficient.
To address this gap, we introduce Titan-Llama, a post-training method for retrofitting attention-based language models with a hierarchical memory architecture that updates parameters at test-time.
Titan-Llama applies a blockwise attention mask and preserves full-fidelity attention over recent tokens while compressing historical context into recurrent, fixed-size neural memory modules (NMMs).
Our formulation allows pre-trained attention layers to be replaced without training a new language model from scratch.
On Llama-3.2-1B-Instruct, Titan-Llama raises accuracy over segmented attention windows from 23.87\% to 40.51\% on QuALITY and from 14.50\% to 40.75\% on LongHealth.
Inference benchmarks show more than 20\% faster prefill, 37\% faster decode at medium context lengths, and 24\% lower steady-state decode memory usage in the evaluated configurations.
Together, these results demonstrate that incorporating memory architecture mechanisms that utilize test-time training effectively modulates long-context behavior of pre-trained models, offering a practical route to adapting pre-trained transformers for efficient long-context inference.
Our code is released on \href{https://github.com/sarahpannn/titans-pytorch/tree/main}{GitHub}.

\end{abstract}

\section{Introduction}

Transformers are efficient to train at scale because attention permits parallel computation across sequence positions.
At inference time, however, retaining keys and values for an increasingly long context cause the KV cache to grow. 
Linear attention and state-space models offer more efficient recurrent inference, but adopting a different architecture generally requires training a new model rather than reusing a pre-trained attention backbone~\citep{katharopoulos2020transformersrnnsfastautoregressive, gu2024mambalineartimesequencemodeling}.
Scaling large language models requires advances in both representational capacity and inference efficiency, which naturally motivates retrofitting the architecture of transformer-based models to better suit them for inference.

The growing KV cache raises the question of how should a pre-trained transformer best retain information from distant tokens without necessarily storing every historical key and value.
Titans~\citep{behrouz2024titanslearningmemorizetest} represents distant information in an RNN dubbed the neural memory module (NMM), effectively parameterizing historical tokens as fast weight updates to a fixed-size NMM.
This motivates our design in using full attention for local windows and representing older tokens in a fixed-size MLP.
It is also worth noting that work in KV cache compaction similarly argues that exacting inference tasks do not necessarily require full fidelity on the KV cache~\citep{zhang2023h2oheavyhitteroracleefficient,zweiger2026fastkvcompactionattention}.

Our work combines these two ideas and develops a method to post-train lightweight, external modules to improve inference efficiency.
We partition attention into local and historical contributions, preserve the local contribution, and train a neural memory module (NMM) to provide information from historical context.
At test time, a self-supervised key-value reconstruction loss updates the NMM as new chunks arrive.
This objective supplies a self-supervised learning signal without labels, although reconstructing key-value pairs does not guarantee exact reproduction of historical attention.
Notably, we keep the pre-trained backbone frozen and are able to recover significant performance while training external weights amounting to only 1\% of the total parameter footprint.

Two main considerations guide our design:
\begin{enumerate}
    \item How can we post-train a pre-trained LLM to use Titans as a means to achieve fixed-size context compaction?
    \item How can we make the resulting model efficient in practice, accounting for the cost of updating memory during prefill as well as storing and reading it during decode?

\end{enumerate}

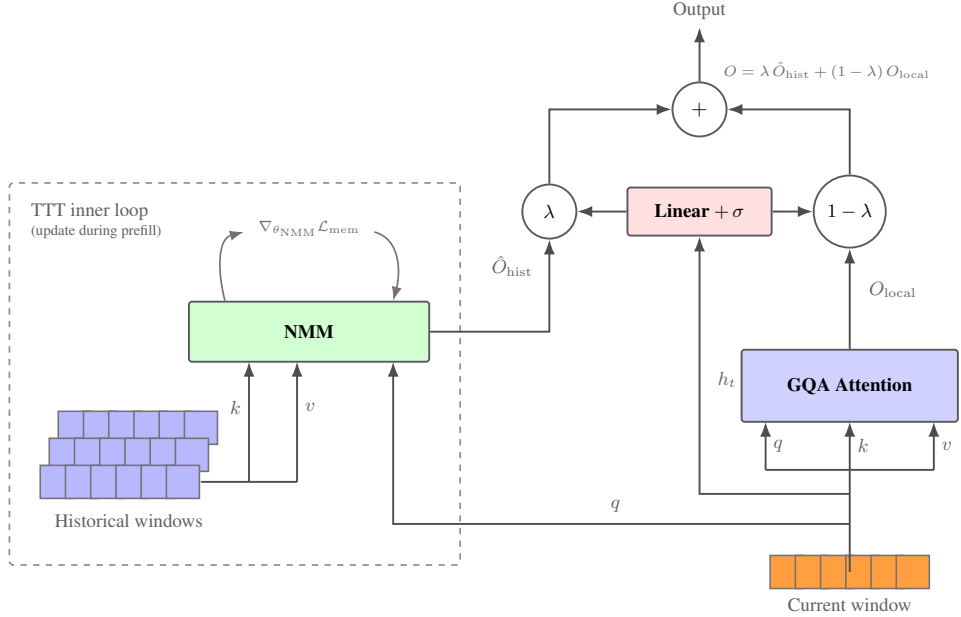
\begin{figure}
    \centering
    \begin{adjustbox}{width=0.95\linewidth}
        \input{titan_llama_fig}
    \end{adjustbox}
    \caption{Titan-Llama architecture at a glance. \textbf{Left:} The inner loss generates gradients at test-time w.r.t. MLP weights, teaching it to memorize historical attention key-value bindings. \textbf{Right:} The query associated with a local hidden state is used to retrieve a value from the NMM. This value, approximating the historical attention output, is then used as a residual and injected into the lambda-gated output stream.}
    \label{fig:schematic}
\end{figure}

\section{Related Work}

\paragraph{Hierarchical Architectures.}
Prior work has found success using hierarchical systems to model long dependencies, or more precisely, different mechanisms to model local and distant relationships~\citep{beltagy2020longformerlongdocumenttransformer, qiu2020blockwiseselfattentionlongdocument, kitaev2020reformerefficienttransformer, bulatov2022recurrentmemorytransformer}.
Many of these approaches, however, do not take advantage of caching and must recompute global attention scores with every new token generated~\citep{patel2024splitwiseefficientgenerativellm, ainslie2020etcencodinglongstructured}.
Instead, we try to design a method that aligns with the abstraction of prefill and decode phases where historical context can be stored in the NMM via test-time training at prefill time.

Other hierarchical approaches apply downsampling operations to the sequence itself, whereas our time steps follow the chronological arrival of the context.~\citep{hwang2025dynamicchunkingendtoendhierarchical,nawrot2022hierarchicaltransformersefficientlanguage,slagle2024spacebytedeletingtokenizationlarge}.

\paragraph{Retrofitting Pre-trained Models.}
Modifications to the attention architecture to leverage transformers' training parallelism to fit inference-time objectives have also been studied in prior literature~\citep{bick2025transformersssmsdistillingquadratic, wang2025mamballamadistillingaccelerating}.
These approaches differ in how much of the original model they preserve and how they train the replacement mechanism.
For instance, some approaches use distillation objectives to fully convert architectures from scratch while others fine-tune modular components~\citep{eyuboglu2025cartridgeslightweightgeneralpurposelong, kojima2025loratttlowranktesttimetraining}.
Our work aims to fine-tune a small, external set of weights that, through training on next token prediction, can learn how to best modulate a frozen backbone.

\paragraph{Test-time Training.}
Test-time training (TTT) methods apply gradient updates to a subset of weights at inference time, creating a meta-objective for the slow weights to be able to ``learn to learn''~\citep{sun2020testtimetrainingselfsupervisiongeneralization, sun2025learninglearntesttime, finn2017modelagnosticmetalearningfastadaptation}.
Recent studies utilize TTT approaches to extend the effective context length or to use it to greater effect~\citep{bansal2025letsnotjustthings}.
However, some methods keep the underlying architecture intact, with gradients applying to the MLP layers, decreasing throughput~\citep{feng2026inplacetesttimetraining}.
The hierarchical approach we use replaces attention for longer-range dependencies with RNNs, which actually improves throughput and average memory footprint for long-context tasks

The TTT works that provide us the most inspiration and are most similar to ours are Titans~\citep{behrouz2024titanslearningmemorizetest} conceptually and LaCT~\citep{zhang2025testtimetrainingright} methodologically, but we focus on the post-training dynamics of retrofitting LLMs rather than pre-training or video.
Importantly, we also use a self-supervised inner objective rather than final logit prediction, as other TTT approaches do~\citep{tandon2025endtoendtesttimetraininglong,kuratov2026gradmemlearningwritecontext, behrouz2025atlaslearningoptimallymemorize}.

\textbf{Titans.} To motivate the attention KV binding used in Titans from first principles, consider the vanilla attention operation, which applies a softmax operation to $\mathbf{q}^\top\mathbf{k}$ before applying it to the $\mathbf{v}$ matrix~\citep{sun2025learninglearntesttime}.
In abstract, we can view this as a two-layer MLP where the softmax serves as an activation function and maps keys to values~\citep{han2026vit3unlockingtesttimetraining, wen2026mixtureoftopkattentionefficientattention}.
In linear attention, by replacing the softmax with a homomorphic function, we collapse the MLP into a single linear map, thereby reducing representational capabilities~\citep{katharopoulos2020transformersrnnsfastautoregressive}.
In the Titans view, efficiency can be achieved by encoding the fast-weight MLP explicitly in neural architecture and simulating fast weights via test-time training (TTT)~\citep{behrouz2024titanslearningmemorizetest, behrouz2025atlaslearningoptimallymemorize}.

\section{Method}
\subsection{Approximating Historical Attention as a Residual} \label{sec:formulation}

As shown in Figure ~\ref{fig:schematic}, our Titan-Llama layers encode distant information in a fixed-size RNN state and local information with attention, loosely resembling the distinction between lossy long-term memories and vivid recent ones.
We now show that contributions to attention from local and historical contexts are easy to partition.
This allows us to approximate full attention by predicting the historical output and the mixing weight.

For a query $\mathbf q$ at position $t$ and segment length $k$, let
  $b_t=k\lfloor t/k\rfloor$ denote the start of the current segment. We partition
  the available keys and values into historical positions
  $\mathcal H=\{i:i<b_t\}$ and local positions
  $\mathcal L=\{i:b_t\leq i\leq t\}$. With scores
  $s_i=\mathbf q^\top\mathbf k_i/\sqrt d$, define
  \[
  Z_{\mathrm{hist}}=\sum_{i\in\mathcal H}e^{s_i},
  \qquad
  Z_{\mathrm{local}}=\sum_{i\in\mathcal L}e^{s_i},
  \]
  and the attention outputs normalized within each partition:
  \[
  O_{\mathrm{hist}}
  =\frac{\sum_{i\in\mathcal H}e^{s_i}\mathbf v_i}{Z_{\mathrm{hist}}},
  \qquad
  O_{\mathrm{local}}
  =\frac{\sum_{i\in\mathcal L}e^{s_i}\mathbf v_i}{Z_{\mathrm{local}}}.
  \]

As in the blockwise formulation of exact attention by ~\citet{dao2022flashattentionfastmemoryefficientexact}, we can partition the softmax numerator and denominator into historical and local contributions, with $\lambda$ representing the fraction of
  attention mass assigned to history:

 \[
  O_{\text{attn}}
  =
  \frac{
  \sum_{i\in\mathcal H} e^{s_i}\mathbf v_i
  +
  \sum_{j\in\mathcal L} e^{s_j}\mathbf v_j
  }{
  \sum_{i\in\mathcal H} e^{s_i}
  +
  \sum_{j\in\mathcal L} e^{s_j}
  },
  \qquad
  \lambda=
  \frac{Z_{\mathrm{hist}}}
  {Z_{\mathrm{hist}}+Z_{\mathrm{local}}}.
  \]

\begin{figure}[H]
    \centering
    \includegraphics[width=0.95\linewidth]{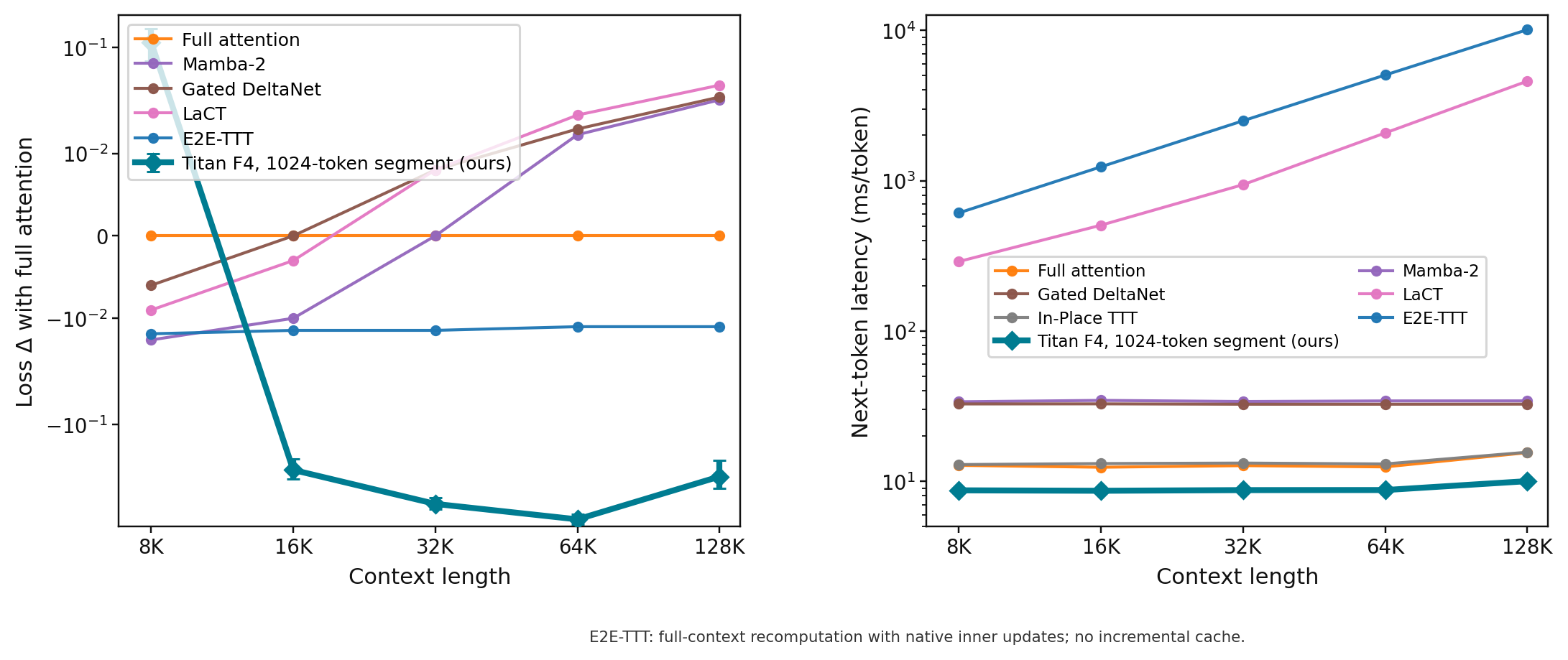}
    \caption{Long-context language modeling and decode latency. \textbf{Left.} Change in per-token negative log-likelihood relative to full attention, measured on Pile.
    Titan-Llama with 1024-token segments beats full attention from 16K through 128K.
  \textbf{Right.} Median single-token decode latency on an A100 at batch size one, after prefill and warmup (lower is better). 
  Titan-Llama's CUDA-graph-enabled decode is faster than the measured full-attention, Mamba-2, Gated DeltaNet, and In-Place TTT implementations across the reported lengths~\citep{dao2024transformersssmsgeneralizedmodels, yang2025gateddeltanetworksimproving}.}
    \label{fig:baseline-loss-delta}
\end{figure}

Finally, we can construct the full attention output as a linear interpolation between the historical and local outputs with parameter $\lambda$:

\begin{equation} \label{eq:derivation}
O
=
\lambda
\underbrace{
\operatorname{softmax}
\left(
\frac{\mathbf q^\top\mathbf K_{\text{hist}}}{\sqrt d}
\right)
\mathbf V_{\text{hist}}
}_{\text{historical attention}}
+
(1-\lambda)
\underbrace{
\operatorname{softmax}
\left(
\frac{\mathbf q^\top\mathbf K_{\text{local}}}{\sqrt d}
\right)
\mathbf V_{\text{local}}
}_{\text{local attention}}
\end{equation}

 \[
  \boxed{
  O=\lambda O_{\text{hist}}+(1-\lambda)O_{\text{local}}.
  }
  \]

Our approach replaces the historical term $O_{\text{hist}}$ with a read $\hat{O}_{\text{hist}}$ from recurrent memory and predicts its mixing weight.
In practice, we predict both $\lambda$ and $O_{\text{hist}}$ and approximate full attention outputs with the expression:  $\hat{\lambda}\hat{O}_{\text{hist}} + (1 - \hat{\lambda})O_{\text{local}}$.
Importantly, this identity does not imply that historical attention can be compressed without loss.

\subsection{Reading and Writing at Test-time} \label{sec:read-write}
Following the derivation in Section \ref{sec:formulation}, we seek to approximate historical attention, or $\hat{O}_{\text{hist}}$, with the external neural memory module (NMM) as used in Titans~\citep{behrouz2024titanslearningmemorizetest}. 
We make the design choice to blend the retrieved output with a learned gate as a residual added to the local attention output, prior to passing the result to the remaining layer computation.
This is similar to other implementations that add the value read from the RNN to the residual stream after attention~\citep{feng2026inplacetesttimetraining} and use it to replace attention entirely in certain layers~\citep{tandon2025endtoendtesttimetraininglong}.

We inject the RNN read as a residual into the attention output because this is a natural implementation of~\eqref{eq:derivation} and aligns with the inner-loss objective described below.
This also allows us to modulate an existing LLM backbone without changing any of the parameters already set.

\textbf{Read.}
As shown in Figure \ref{fig:schematic}, we use the NMM to memorize the attention key–value mappings from historical attention windows and provide a residual to the current local-attention output, with the two contributions balanced by a per-channel $\lambda$ vector.
More specifically, for each query $\mathbf{q}$, the NMM retrieves a historical attention residual $\hat{\mathbf{O}}_{\text{hist}}=\mathcal{M}(\mathbf{q})$, while a learned gate is computed per channel as:

\[
\mathbf{\hat{\lambda}}=\sigma\left(\mathbf{W}_{\lambda}^{\top}\operatorname{RMSNorm}(\mathbf{q})+\mathbf{b}_{\lambda}\right).
\]
We combine this local attention output with the NMM retrieval as $\hat{\mathbf{O}}_{\text{full attention}} = \hat{\lambda} \hat{\mathbf{O}}_{\text{hist}} + (1 - \hat{\lambda}) \mathbf{O}_{\text{local}}$.
We show this expression approximates $\mathbf{O}_{\text{full attention}}$ in Section \ref{sec:formulation}.
For tokens in a chunk, the NMM read uses the state accumulated from preceding chunks.
During inference, once a full chunk has been processed, its keys and values are used to compute the loss that updates the NMM.
    
\textbf{Write.}
To update the NMM at inference time, we minimize a squared-error key–value reconstruction loss over each multi-token chunk $c$, binding a map between $\mathbf{K}$ and $\mathbf{V}$:
$$
\mathcal{L}_c=\left\|\mathcal{M}(\mathbf{K}_c)-\mathbf{V}_{\text{c}}\right\|^2.
$$

This loss gives the memory module a training signal at every chunk without requiring labels or a teacher attention output. 
It encourages the module to retain associations present in the historical keys and values, which future queries may use for retrieval. 
A future query, however, need not coincide with a stored key, and historical attention can combine values from several positions.
Key--value reconstruction is therefore a surrogate for useful memory, rather than an exact objective for reproducing historical attention.

Optimization is done via gradient descent on the NMM once per chunk using learned momentum, decay, and step-size parameters:

$$
\mathcal{M}_c=(1-\alpha_c)\mathcal{M}_{c-1}+\mathbf{S}_c,\qquad
\mathbf{S}_c=\eta_c\mathbf{S}_{c-1}-\theta_c\nabla_{\mathbf{M}}\mathcal{L}_c(\mathcal{M}_{c-1}).
$$

In practice, we do not perform this update on a per-token scale, and instead use fixed-size token batches to take advantage of modern hardware.
For even more efficiency, we use an offline variant of this algorithm by batching the gradient update over multiple chunks.

\subsection{Architecture}
We make several architectural choices to adapt pretrained LLMs for efficient inference.
First, blockwise attention allows us to decouple long-term relationships within the context.
This mask complicates the interpretation of~\eqref{eq:derivation} because earlier layers also use segmented attention, so the historical keys and values differ from those produced by the full-attention model.
Blockwise attention also provides a natural threshold for NMM hidden-state updates and inference-time gains from sparse computation.
    
As shown in \eqref{eq:derivation}, attention outputs are the mass-weighted sum of local and historical contributions, which we aim to simulate via a predicted $\lambda$.
This is nontrivial as our use of $\lambda$ essentially substitutes the usage of a softmax normalization as a means of token reweighting and mixing.
We choose to utilize a linear layer on top of the hidden input stream to predict $\hat{\lambda}$ as it is a query-dependent quantity, and we use it to interpolate between $O_{\text{local}}$ and $\hat{O}_{\text{hist}}$.

\begin{figure}[H]
    \centering
    \includegraphics[width=1.0\linewidth]{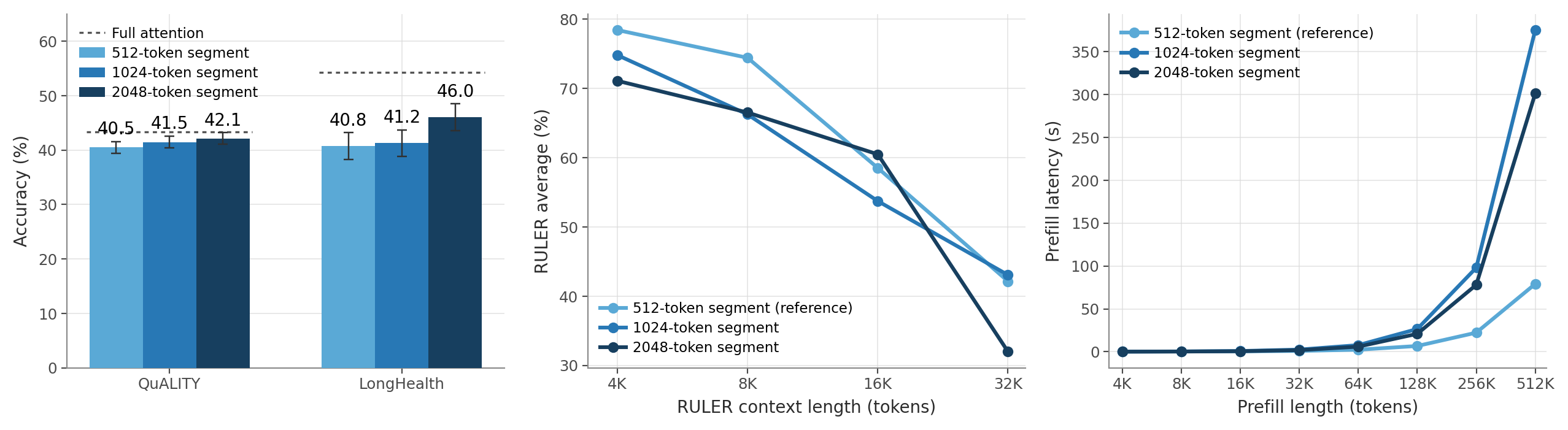}
    \caption{The effects of scaling chunk size on downstream performance and efficiency. \textbf{Left.} Scaling chunk size improves performance on long-context reasoning tasks like QuALITY and LongHealth. \textbf{Middle.} RULER trends are mixed. Longer chunks help at intermediate lengths but do not consistently improve associative recall. Detailed RULER results can be found in Table \ref{tab:ruler-grouped} in the Appendix. \textbf{Right.} Larger chunks increase prefill latency as memory bandwidth saturates.}
    \label{fig:scaling-cs}
\end{figure}

\section{Experiments} \label{sec:experiments}
We convert the first four layers of Llama-3.2-1B-Instruct ~\citep{grattafiori2024llama3herdmodels} to blockwise attention and add a neural memory module for each KV head.
The remaining layers retain full attention. 
This setting tests whether memory can recover information no longer accessible through attention in the converted layers, without requiring every layer to rely on the new mechanism. 

Each three-layer NMM learns historical key-value bindings using a self-supervised test-time objective.
During post-training, we optimize the external modules for language modeling while keeping the pre-trained backbone frozen.
We conduct a series of experiments to measure how much long-text performance the NMM recovers, which forms of recall it supports, and how our architectural design affects inference costs.


\subsection{Post-Training for Long Context Reasoning}
We post-train Titan-Llama on long sequences using a standard next-token prediction loss, while the NMM continues to update its fast weights through the self-supervised key-value objective in Section~\ref{sec:read-write}.
The language modeling loss trains the added components to make those memory reads useful to the frozen backbone.
We initially considered distilling the backbone's historical attention output into the NMM, but found this difficult to scale and ineffective in practice.
Under our final objective, $\hat{O}_{\text{hist}}$ need not reproduce that output exactly and instead learns a contribution that helps predict future tokens under segmented attention. 


We show that training Titan-Llama on next-token prediction, the details of which can be found in Appendix \ref{sec:training-details}, improves not just next-token prediction but also recall and reasoning over long-term dependencies.
We evaluate long-context question answering on QuALITY ~\citep{pang2022qualityquestionansweringlong} and LongHealth~\citep{adams2024longhealthquestionansweringbenchmark}.
Results are shown in Figure ~\ref{fig:scaling-cs} and compared against a segmented attention baseline to show the effect induced by adding the NMM.

Figure~\ref{fig:baseline-loss-delta} shows how language-modeling loss and decode latency change with context length. 
Loss delta is the difference in per-token negative log-likelihood from full-attention Llama, so values below zero indicate better performance. 
With 1024-token segments, Titan-Llama stays below zero from 16K through 128K tokens, while the measured Mamba-2, Gated DeltaNet, and LaCT baselines incur larger loss penalties at longer contexts. 
Because the model is training on sequences of at most 16K tokens, the measurements beyond 16K also test how its language-modeling behavior extends beyond the training length.


\begin{table}[t]
    \centering
    \scriptsize
    \setlength{\tabcolsep}{1.5pt}
     \caption{RULER S-NIAH accuracy (\%) across context lengths.
  Titan-Llama matches full attention on 32K pass-key retrieval and achieves
  the strongest non-full-attention UUID scores at 8K--32K.
  Relative decode time ($\downarrow$) averages next-token latency divided by
  full-attention latency at 8K, 16K, and 32K on A100.
  $\dagger$ marks timings of publicly available code that may not utilize caching.}
    \label{tab:ruler-niah-baselines}
    \resizebox{\linewidth}{!}{%
    \begin{tabular}{lccc ccc ccc c}
      \toprule
      & \multicolumn{3}{c}{S-NIAH-1 (pass-key)}
      & \multicolumn{3}{c}{S-NIAH-2 (number)}
      & \multicolumn{3}{c}{S-NIAH-3 (UUID)}
      & \shortstack{Relative decode\\time $\downarrow$} \\
      \cmidrule(lr){2-4} \cmidrule(lr){5-7}
      \cmidrule(lr){8-10} \cmidrule(l){11-11}
      Model & 8K & 16K & 32K & 8K & 16K & 32K
            & 8K & 16K & 32K & Mean \\
      \midrule
      Full attention & \textbf{100.0} & \textbf{100.0} & \textbf{100.0} &
                       99.0 & \textbf{100.0} & \textbf{100.0} &
                       64.0 & 64.0 & \textbf{83.0} & 1.00 \\
      \midrule
      Mamba-2        & 99.0 & 49.0 & 26.0 &
                       99.0 & 43.0 & 28.0 &
                       77.0 & 36.0 & 24.0 & 2.70 \\
      Gated DeltaNet & \textbf{100.0} & 50.0 & 26.0 &
                       \textbf{100.0} & 43.0 & 27.0 &
                       91.0 & 45.0 & 23.0 & 2.59 \\
      LaCT           & 98.0 & 43.0 & 22.0 &
                       \textbf{100.0} & 43.0 & 27.0 &
                       74.0 & 34.0 & 23.0 & $45.8^{\dagger}$ \\
      TTT-E2E        & \textbf{100.0} & 46.0 & 24.0 &
                       99.0 & 43.0 & 28.0 &
                       77.0 & 44.0 & 24.0 & $115^{\dagger}$ \\
      \midrule
      Titan-Llama    & \textbf{100.0} & \textbf{100.0} & \textbf{100.0} &
                       96.0 & 78.0 & 88.0 &
                       \textbf{92.0} & \textbf{86.0} & 48.0 & \textbf{0.689} \\
      \bottomrule
    \end{tabular}%
    }
  \end{table}

\subsection{RULER Evaluation}
Aggregate language-modeling loss does not tell us which kinds of historical information are best encoded by the NMM.
The RULER benchmark measures different types of recall over long sequences, including retrieval of individual values, multiple associations, and word-frequency information~\citep{hsieh2024rulerwhatsrealcontext}.
The RULER results in Table \ref{tab:ruler} compare full attention, segmented attention without memory, and Titan-Llama.
The NMMs are able to effectively enrich representations enough to recover the majority of backbone performance, particularly for single-needle and multi-key retrieval at lengths up to 16K.

Unexpectedly, Titan-Llama also exceeds the full-attention baseline on multi-key retrieval at 4K–16K and on frequent/common word extraction at 8K–32K.
Multi-value retrieval remains below full attention for every length, suggesting greater difficulty returning several distinct values from the same context.
Performance declines at 32K, beyond the 16K training length, suggesting limited length generalization in this setting.
This reveals that the NMM can perform semantic retrieval, accumulation, and recall of individual values, but is less proficient at recovering several values at once.

When compared to other efficient architectures in Table \ref{tab:ruler-niah-baselines}, Titan-Llama matches the full attention baseline on pass-key retrieval and exceeds full attention performance on UUID recall at 8K and 16K.
On 32K pass-key retrieval, it also retains full accuracy where the measured efficient-model baselines decline. These comparisons show that compacting history can preserve strong retrieval on some tasks, even though the broader RULER results show that it does not preserve every form of recall equally well.
Especially compared to RNN-based models with limited full-attention layers such as Mamba-2 and Gated DeltaNet, Titan-Llama's performance does not degrade as severely with increasing context lengths.

\begin{table}[t]
    \caption{RULER accuracy by task family and context length. Titan-Llama recovers much of the accuracy lost to attention segmentation, although the gap to full attention widens on some tasks at longer contexts. Multi-key and FWE/CWE results exceed the full attention baseline.} \label{tab:ruler}
    \centering
    \small
    \begin{tabular}{lcccccccccccc}                                                       
    \toprule
    Model & \multicolumn{4}{c}{Single needle} & \multicolumn{4}{c}{Multi-key} &
  \multicolumn{4}{c}{Multi-query} \\
    \cmidrule(lr){2-5} \cmidrule(lr){6-9} \cmidrule(lr){10-13}
     & 4K & 8K & 16K & 32K & 4K & 8K & 16K & 32K & 4K & 8K & 16K & 32K \\
    \midrule
    Full attention & \textbf{100.0} & \textbf{99.3} & \textbf{100.0} & \textbf{99.3} & 82.0 &
  74.0 & 63.3 & \textbf{57.3} & \textbf{99.5} & \textbf{94.0} & \textbf{94.0} & \textbf{83.5}
  \\
    Segmented attn. & 10.7 & 0.0 & 0.0 & 0.0 & 0.0 & 0.0 & 0.0 & 0.0 & 0.0 & 0.0 & 0.0 & 0.0
  \\
    Titan-Llama & 97.3 & 96.0 & 88.0 & 78.7 & \textbf{91.3} & \textbf{88.0} & \textbf{66.7} &
  39.3 & 96.5 & 90.5 & 70.5 & 34.5 \\
    \midrule[\heavyrulewidth]
    Model & \multicolumn{4}{c}{Multi-value} & \multicolumn{4}{c}{FWE/CWE} &
  \multicolumn{4}{c}{HotpotQA} \\
    \cmidrule(lr){2-5} \cmidrule(lr){6-9} \cmidrule(lr){10-13}
     & 4k & 8k & 16k & 32k & 4k & 8k & 16k & 32k & 4k & 8k & 16k & 32k \\
    \midrule                                                                             
    Full attention & \textbf{99.5} & \textbf{97.5} & \textbf{89.0} & \textbf{70.0} &
  \textbf{59.1} & 39.2 & 22.1 & 20.0 & \textbf{48.0} & 40.0 & \textbf{46.0} & \textbf{36.0} \\
    Segmented attn. & 1.5 & 0.0 & 0.0 & 0.0 & 27.0 & 9.8 & 3.3 & 0.0 & 8.0 & 0.0 & 0.0 & 0.0
  \\
    Titan-Llama & 88.0 & 81.0 & 63.0 & 53.0 & 53.4 & \textbf{44.9} & \textbf{32.7} &
  \textbf{24.8} & 40.0 & \textbf{46.0} & 28.0 & 26.0 \\
    \bottomrule
    \end{tabular}
\end{table}

\subsection{Inference Gains}
While various studies have applied TTT to modulate existing architecture in transformer, full attention oftentimes serves as a lower bound on performance and a ceiling on inference efficiency.~\citep{feng2026inplacetesttimetraining}
Because our use of TTT adapts a cheaper replacement to long-context relationships, we are able to harness greater inference efficiency.
The decode comparison in Figure~\ref{fig:baseline-loss-delta} shows this in effect as Titan-Llama has lower median single-token latency than the measured full-attention and efficient-model baselines across the reported context lengths.

\begin{figure}[t]
\centering
\includegraphics[width=1.0\linewidth]{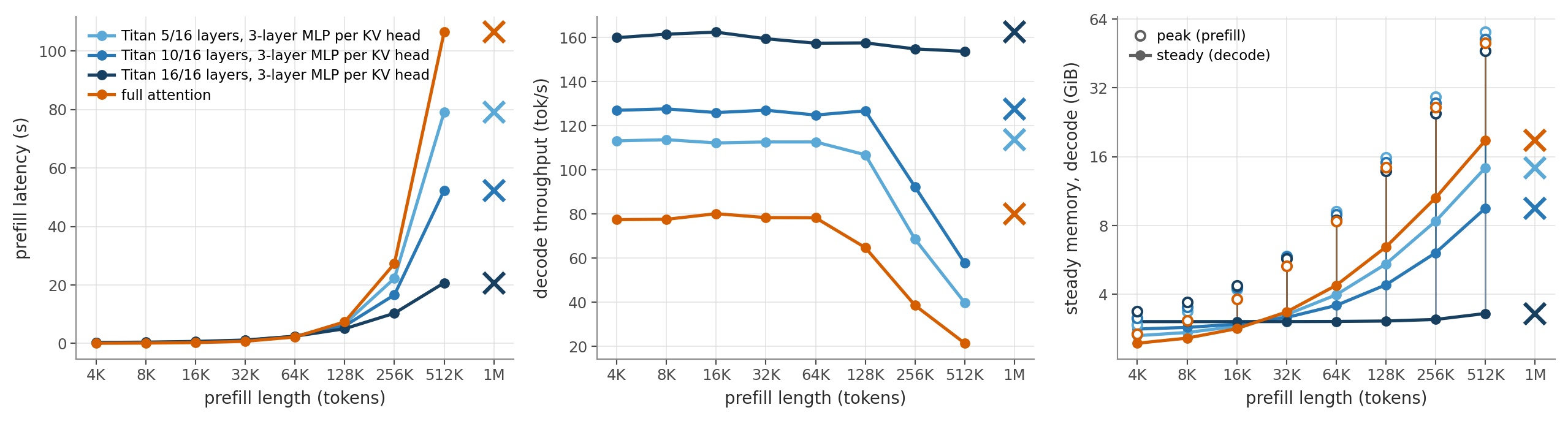}
\caption{Increasing the proportion of Llama 3.2 1B layers retrofitted with NMMs increases token decode throughput, decreases prefill latency, and also decreases memory footprint during token decoding. Peak memory footprint during prefill for partial Titan-Llama hybrids exceeds full-attention Llama while retrofitting all layers leads to lower peak memory footprint. Benchmarking was performed on a single 80GB NVIDIA A100 GPU.}
\label{fig:speed}
\end{figure}

Figure~\ref{fig:speed} separately measures how these costs change as we convert more layers. 
At longer prompt lengths, Titan-Llama achieves lower prefill latency than the full-attention backbone, and the advantage grows with the number of converted layers, the details of which can be found in Appendix~\ref{sec:inference-implementation}.
In fact, the prefill process for Titan-Llama is more than 20\% faster than full-attention at long prefill lengths with only five converted layers.
The steady-state decode memory usage for Titan-Llama is also 24\% lower than full attention with the KV cache, though peak memory usage during prefill is more expensive for partial hybrids.
This reflects our intuition that our approach should result in inference economics gains, with one practical limitation that the \textit{peak} memory footprint of Titan-Llama exceeds that of full attention in shorter sequence lengths and in hybrid variants.

\subsection{Scaling chunk size} \label{sec:scaling}
The chunk size in Titan-Llama corresponds to the number of tokens in a local window as well as the effective token batch size for the inner objective loss. 
A similar hyperparameter affects not just inference dynamics but capabilities in other studies as well~\cite{feng2026inplacetesttimetraining}.
Changing it affects both what information remains available through exact local attention and how often the NMM is updated.
The RULER and long-context reasoning results seen previously utilized 512-token segments.
Figure \ref{fig:scaling-cs} varies the chunk size from 512 to 2048 tokens.
Larger chunks improve accuracy on QuALITY and LongHealth, but their effects on RULER are mixed: they help some task–length combinations while reducing accuracy on others, including several needle retrieval settings.

We also study the effects of scaling context length on RULER performance and show in Appendix Table~\ref{tab:ruler-grouped} that shorter context lengths lead to better performance on NIAH-style tasks while longer chunk sizes lend better semantic retrieval.

Larger chunks also increase measured prefill latency. Chunk size thus sets a task-dependent balance among local attention, the frequency of memory updates, and inference cost, rather than providing a uniform accuracy gain.




\section{Discussion}
The idea of using test-time training on an MLP to improve language model capabilities, whether on an existing part of LLM architecture or on external modules, has been well explored in a substantial body of work.~\citep{behrouz2024titanslearningmemorizetest, feng2026inplacetesttimetraining, tandon2025endtoendtesttimetraininglong, sun2025learninglearntesttime, zhang2025testtimetrainingright}
Compared to these studies, however, our main contribution seeks to take advantage of the theoretical understanding of the updated set of weights as a memory mechanism and use it to \textit{substitute} the long-range attention calculations of a pre-trained backbone.

To that end, our results corroborate the current line of research which utilizes TTT to improve language modeling, while demonstrating practical inference-efficiency benefits.
Similar to~\citet{zhang2025testtimetrainingright, behrouz2024titanslearningmemorizetest}, we show that TTT using KV binding effectively learns to do language modeling.
And like~\citet{feng2026inplacetesttimetraining}, we show that TTT can extend the context window for certain tasks. 

Our work also allows for open questions such as scaling NMM size or changing the NMM architecture.
Further, understanding the mechanisms behind memory retrieval is a natural extension of this work, and exploring the use of inner-loop objectives other than KV binding could lead to improvements on this method.

\subsubsection*{Acknowledgments}
We would like to extend our appreciation to the MIT Office of Research Computing and Data for providing the computational resources to run our experiments. We would also like to extend thanks to Sarah Dufays for her help in early versions of this project.

\bibliography{iclr2027_conference}
\bibliographystyle{iclr2027_conference}

\appendix

  \section{Training Details}
  \label{sec:training-details}

  \subsection{Datasets}

  We train on one billion tokens of packed sequences from FineWeb, with a maximum sequence length of 16K
  tokens~\citep{penedo2024finewebdatasetsdecantingweb}. We then perform supervised fine-tuning on
  LongAlign~\citep{bai2024longalignrecipelongcontext} and LoongRL~\citep{wang2025loongrlreinforcementlearningadvanced}. A final fine-tuning stage uses the QuALITY training set to adapt the model to multiple-choice
  answering~\citep{pang2022qualityquestionansweringlong}.

  \subsection{Hyperparameters}

  The first four Llama layers each contain one neural memory module (NMM) per KV head. Each NMM is a three-layer network with a hidden
  dimension of 64 and GELU activations. The added components contain 17.23M trainable parameters in total.

  We use a batch size of 64 during the initial training stage, corresponding to approximately one million tokens per batch at the
  maximum sequence length. Supervised fine-tuning uses a batch size of 128 with variable answer lengths. The inner-loss batch size
  equals the chunk size. We evaluate chunk sizes of 512, 1024, and 2048 tokens in Section~\ref{sec:scaling}. Compute constraints
  prevented a broad search over these architectural and training choices.

   \begin{figure}[H]
      \centering
      \includegraphics[width=\linewidth]{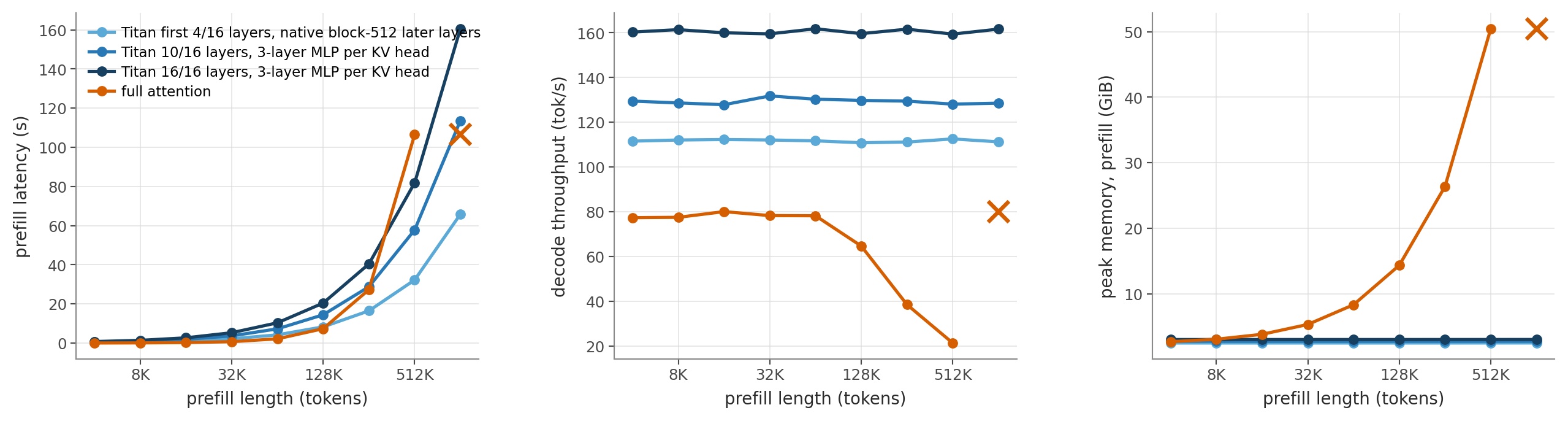}
      \caption{Inference cost when later attention layers are restricted to local windows. The plot shows a 512-token window.
      Restricting more layers limits the growth of the KV cache and makes decode throughput less sensitive to prefill length.}
      \label{fig:abl-speed}
  \end{figure}

  \section{Inference Implementation} \label{sec:inference-implementation}

  Autoregressive decoding performs many small operations for each generated token. At batch size one, the cost of launching those
  operations can be substantial. For the measurements in Figure~\ref{fig:speed}, we compile the Titan-Llama layers in CUDA Graph
  ``reduce-overhead'' mode and replay their single-token decode computations. Compilation and graph capture take place during warmup,
  outside the timed decode steps.

  The captured computation uses a preallocated KV ring buffer for each attention segment. Cache positions and memory-store indices are
  updated in place on the GPU, and attention operands retain fixed shapes. We capture separate paths for ordinary tokens and for tokens
  that complete a memory-update chunk. The remaining full-attention layers are compiled without graph capture so that they retain their
  optimized attention implementation.

  CUDA Graphs reduce launch overhead but still execute the underlying kernels. A fused implementation of the memory read, update, and
  output blend could reduce intermediate memory traffic and launch overhead further. Such an implementation would need to preserve the
  ordering of reads and updates at chunk boundaries.

  \section{Additional Experiments}

  \begin{table}[t]
      \caption{Multiple-choice accuracy when the later attention layers are also restricted to local windows. Neural memory improves
      both tasks relative to segmented attention alone, but does not recover the full-attention baseline.}
      \label{tab:abl-long-context-mc}
      \centering
      \begin{tabular}{lcc}
          \toprule
          \textbf{Model} & \textbf{QuALITY} & \textbf{LongHealth} \\
          \midrule
          Full attention (Llama-3.2-1B-Instruct)
              & $\mathbf{43.34} \pm 1.08$ & $\mathbf{54.25} \pm 2.49$ \\
          \midrule
          \multicolumn{3}{l}{\emph{Local-visibility variants}} \\
          \quad Segmented attention, no NMM & 26.8 & 15.8 \\
          \quad Segmented attention, with NMM & 35.2 & 32.3 \\
          \bottomrule
      \end{tabular}
  \end{table}

  \subsection{Restricting the Visibility of Later Attention Layers}
  \label{sec:local-visibility}

  In the main configuration, the layers after the Titan-Llama layers retain full attention. They can therefore access earlier tokens
  directly, even though the converted layers use segmented attention and neural memory. To test how much the model relies on this
  remaining full attention, we also restrict the later layers to local windows. In this variant, neural memory is the only route by
  which information from a completed segment can affect later segments.

  The restriction reduces inference cost, as shown in Figure~\ref{fig:abl-speed}, but it also causes a large loss in accuracy. 
  On  QuALITY, adding the NMM raises accuracy from 26.8 to 35.2; on LongHealth, it raises accuracy from 15.8 to 32.3 (Table~\ref{tab:abl-long-context-mc}). 
  Both results remain below the full-attention baseline. The RULER results in Table~\ref{tab:abl-ruler} show the
  same limitation more sharply: the NMM provides occasional gains over segmented attention, but neither configuration reliably retrieves information across segments.

  \begin{table}[h]
      \caption{RULER accuracy by task family and context length when later attention layers are restricted to local windows. Results
      use cached autoregressive decoding. Accuracy remains low on most tasks, indicating that the NMM does not recover the information
      available through the later full-attention layers in the main configuration.}
      \label{tab:abl-ruler}
      \centering
      \small
      \setlength{\tabcolsep}{2pt}
      \resizebox{\textwidth}{!}{%
      \begin{tabular}{l*{16}{c}}
          \toprule
          Model
          & \multicolumn{4}{c}{Single needle}
          & \multicolumn{4}{c}{Multi-key}
          & \multicolumn{4}{c}{Multi-query}
          & \multicolumn{4}{c}{Multi-value} \\
          \cmidrule(lr){2-5}
          \cmidrule(lr){6-9}
          \cmidrule(lr){10-13}
          \cmidrule(lr){14-17}
          & 4k & 8k & 16k & 32k
          & 4k & 8k & 16k & 32k
          & 4k & 8k & 16k & 32k
          & 4k & 8k & 16k & 32k \\
          \midrule
          Segmented attn.
          & 0.0 & 3.3 & 1.3 & 0.7
          & 0.3 & 1.3 & 0.0 & 0.0
          & 2.0 & 1.8 & 0.8 & 0.3
          & 1.0 & 3.8 & 2.0 & 0.0 \\
          Titan-Llama
          & 1.0 & 4.7 & 4.0 & 1.7
          & 1.3 & 2.3 & 0.7 & 1.0
          & 3.5 & 2.8 & 1.0 & 1.3
          & 1.0 & 3.8 & 1.8 & 1.0 \\
          \midrule[\heavyrulewidth]
          Model
          & \multicolumn{4}{c}{FWE/CWE}
          & \multicolumn{4}{c}{HotpotQA}
          & \multicolumn{4}{c}{SQuAD}
          & \multicolumn{4}{c}{VT} \\
          \cmidrule(lr){2-5}
          \cmidrule(lr){6-9}
          \cmidrule(lr){10-13}
          \cmidrule(lr){14-17}
          & 4k & 8k & 16k & 32k
          & 4k & 8k & 16k & 32k
          & 4k & 8k & 16k & 32k
          & 4k & 8k & 16k & 32k \\
          \midrule
          Segmented attn.
          & 22.7 & 10.1 & 4.2 & 2.7
          & 17.0 & 9.0 & 1.0 & 2.0
          & 31.4 & 1.0 & 0.0 & 0.0
          & 0.4 & 0.0 & 0.0 & 0.0 \\
          Titan-Llama
          & 23.6 & 8.1 & 3.8 & 2.7
          & 20.0 & 8.0 & 11.0 & 16.0
          & 27.3 & 4.0 & 2.3 & 3.0
          & 0.0 & 0.0 & 0.0 & 0.0 \\
          \bottomrule
      \end{tabular}%
      }
  \end{table}

  \subsection{Chunk Size and RULER Performance}

  Chunk size controls both the span of local attention and the number of tokens used for each memory update. Table~\ref{tab:ruler-grouped} reports RULER results for 512-, 1024-, and 2048-token chunks. 
  The effect depends on the task and context length. 
  The 512- token setting is strongest on most multi-key and multi-value retrieval settings, while larger chunks improve some individual results, including 16K SQuAD and multi-value accuracy. 
  These results do not support a single best chunk size across the RULER tasks.

  \begin{table}[h]
      \caption{RULER accuracy by task family, context length, and chunk size. Bold marks the best result among the three chunk sizes
      for each task and length.}
      \label{tab:ruler-grouped}
      \centering
      \small
      \setlength{\tabcolsep}{2pt}
      \resizebox{\textwidth}{!}{%
      \begin{tabular}{l*{16}{c}}
          \toprule
          Chunk size
          & \multicolumn{4}{c}{Single needle}
          & \multicolumn{4}{c}{Multi-key}
          & \multicolumn{4}{c}{Multi-query}
          & \multicolumn{4}{c}{Multi-value} \\
          \cmidrule(lr){2-5}
          \cmidrule(lr){6-9}
          \cmidrule(lr){10-13}
          \cmidrule(lr){14-17}
          & 4k & 8k & 16k & 32k
          & 4k & 8k & 16k & 32k
          & 4k & 8k & 16k & 32k
          & 4k & 8k & 16k & 32k \\
          \midrule
          512
          & \textbf{97.3} & \textbf{96.0} & 88.0 & \textbf{78.7}
          & \textbf{91.3} & \textbf{88.0} & 66.7 & \textbf{39.3}
          & \textbf{96.5} & \textbf{90.5} & 70.5 & 34.5
          & \textbf{88.0} & \textbf{81.0} & 63.0 & \textbf{53.0} \\
          1024
          & 95.3 & 92.7 & 91.0 & \textbf{78.7}
          & 81.3 & 79.3 & 63.0 & 36.3
          & 92.8 & 85.0 & 51.2 & \textbf{42.2}
          & 87.0 & 68.8 & 49.0 & 38.2 \\
          2048
          & 90.3 & 92.7 & \textbf{91.7} & 51.7
          & 66.3 & 74.0 & \textbf{67.0} & 32.3
          & 91.3 & 88.8 & \textbf{81.2} & 19.2
          & 87.5 & 79.2 & \textbf{82.2} & 25.0 \\
          \midrule[\heavyrulewidth]
          Chunk size
          & \multicolumn{4}{c}{FWE/CWE}
          & \multicolumn{4}{c}{HotpotQA}
          & \multicolumn{4}{c}{SQuAD}
          & \multicolumn{4}{c}{VT} \\
          \cmidrule(lr){2-5}
          \cmidrule(lr){6-9}
          \cmidrule(lr){10-13}
          \cmidrule(lr){14-17}
          & 4k & 8k & 16k & 32k
          & 4k & 8k & 16k & 32k
          & 4k & 8k & 16k & 32k
          & 4k & 8k & 16k & 32k \\
          \midrule
          512
          & 53.4 & \textbf{44.9} & 32.7 & \textbf{24.8}
          & 40.0 & \textbf{46.0} & 28.0 & 26.0
          & 55.5 & 40.2 & \textbf{36.2} & 21.5
          & \textbf{66.8} & \textbf{68.0} & \textbf{33.6} & 9.2 \\
          1024
          & 54.0 & 38.7 & 26.5 & 21.0
          & \textbf{43.0} & 37.0 & 34.0 & \textbf{29.0}
          & 62.4 & 39.4 & 32.1 & \textbf{31.0}
          & 49.4 & 37.6 & 17.0 & \textbf{32.2} \\
          2048
          & \textbf{55.1} & 36.5 & \textbf{32.9} & 22.4
          & 40.0 & 33.0 & \textbf{36.0} & \textbf{29.0}
          & \textbf{67.1} & \textbf{44.1} & 32.8 & 27.7
          & 58.2 & 46.6 & 12.2 & 18.0 \\
          \bottomrule
      \end{tabular}%
      }
  \end{table}



\end{document}

%% file: math_commands.tex
\usepackage{amsmath,amsfonts,bm}

\def\eqref#1{equation~\ref{#1}}

\def\1{\bm{1}}

\DeclareMathAlphabet{\mathsfit}{\encodingdefault}{\sfdefault}{m}{sl}
\SetMathAlphabet{\mathsfit}{bold}{\encodingdefault}{\sfdefault}{bx}{n}



%% file: titan_llama_fig.tex
\usetikzlibrary{arrows.meta,calc,fit}

\begin{tikzpicture}[
    font=\small,
    >={Latex[length=2mm,width=1.4mm]},
    flow/.style={draw=black!70, line width=0.9pt, -Latex},
    bus/.style={draw=black!70, line width=0.9pt},
    grad/.style={draw=black!55, line width=0.8pt, -Latex},
    module/.style={draw=black!65, rounded corners=1.5pt, line width=0.9pt,
                   minimum height=10mm, align=center, fill=white},
    nmm/.style={module, fill=green!18, minimum width=40mm},
    attn/.style={module, fill=blue!18, minimum width=36mm, minimum height=12mm},
    gatebox/.style={module, fill=red!12, minimum width=24mm, minimum height=8mm},
    gate/.style={draw=black!65, circle, minimum size=9mm, inner sep=1pt,
                 line width=0.9pt, fill=white},
    plus/.style={gate, font=\bfseries},
    token/.style={draw=black!55, line width=0.75pt,
                  minimum width=5.5mm, minimum height=5.5mm},
    histtok/.style={token, fill=blue!28},
    currtok/.style={token, fill=orange!75},
    ann/.style={text=black!62},
    smallann/.style={text=black!58, font=\scriptsize},
    dashedbox/.style={draw=black!45, dashed, rounded corners=2pt, line width=0.8pt}
]

\begin{scope}[shift={(-8.2,-3.5)}]
    \foreach \i in {0,...,5} \node[histtok] (hb\i) at ({0.42*\i+0.30},0.90) {};
    \foreach \i in {0,...,5} \node[histtok] (hm\i) at ({0.42*\i+0.15},0.45) {};
    \foreach \i in {0,...,5} \node[histtok] (hf\i) at ({0.42*\i},0.00) {};
\end{scope}
\node[ann] (histlab) at (-7.0,-4.15) {Historical windows};

\node[nmm] (nmm) at (-4.0,-1.0) {\bfseries NMM};

\node[smallann] (gradlab) at (-4.0,0.72)
    {$\nabla_{\theta_{\mathrm{NMM}}}\mathcal{L}_{\mathrm{mem}}$};
\draw[grad] ($(nmm.north)+(-1.4,0)$)
    to[out=105,in=195] ($(gradlab.west)+(-0.12,-0.06)$);
\draw[grad] ($(gradlab.east)+(0.12,-0.06)$)
    to[out=-15,in=75] ($(nmm.north)+(1.4,0)$);

\node[dashedbox, fit=(hf0)(hb5)(nmm)(gradlab)(histlab), inner sep=5mm] (tttbox) {};
\node[ann, anchor=north west, align=left] at ($(tttbox.north west)+(0.25,-0.22)$)
    {TTT inner loop\\[-1pt]{\scriptsize (update during prefill)}};

\draw[bus] (-5.80,-3.5) -- (-4.20,-3.5);
\draw[flow] (-5.00,-3.5) -- (-5.00,-1.5) node[pos=0.62,left,ann] {$k$};
\draw[flow] (-4.20,-3.5) -- (-4.20,-1.5) node[pos=0.62,right,ann] {$v$};

\begin{scope}[shift={(3.95,-5.0)}]
    \foreach \i in {0,...,5} \node[currtok] (c\i) at ({0.42*\i},0) {};
\end{scope}
\node[ann] at (5.0,-5.55) {Current window};

\node[attn] (gqa) at (5.0,-1.9) {\bfseries GQA Attention};

\draw[bus] (5.0,-5.0) -- (5.0,-3.3);
\draw[bus] (3.6,-3.3) -- (6.4,-3.3);
\draw[flow] (3.6,-3.3) -- (3.6,-2.5) node[pos=0.5,right,ann] {$q$};
\draw[flow] (5.0,-3.3) -- (5.0,-2.5) node[pos=0.5,right,ann] {$k$};
\draw[flow] (6.4,-3.3) -- (6.4,-2.5) node[pos=0.5,right,ann] {$v$};

\node[gate] (lam) at (0,1.0) {$\lambda$};
\node[gate, minimum size=12mm] (loc) at (5.0,1.0) {$1-\lambda$};
\node[gatebox] (lin) at (2.5,1.0) {\bfseries Linear $+\,\sigma$};

\draw[flow] (lin.west) -- (lam.east);
\draw[flow] (lin.east) -- (loc.west);

\draw[flow] (5.0,-3.70) -- (2.5,-3.70) -- (lin.south);
\node[ann, anchor=west] at (2.68,-1.80) {$h_t$};

\draw[flow] (5.0,-4.20) -- (-2.6,-4.20) -- (-2.6,-1.5);
\node[ann, anchor=south] at (1.1,-4.12) {$q$};

\draw[flow] (nmm.east) -- (0,-1.0) -- (lam.south);
\node[ann, anchor=east] at (-0.15,0.10) {$\hat O_{\mathrm{hist}}$};

\draw[flow] (gqa.north) -- (loc.south);
\node[ann, anchor=west] at (5.20,-0.30) {$O_{\mathrm{local}}$};

\node[plus] (sum) at (2.5,2.7) {$+$};
\draw[flow] (lam.north) |- (sum.west);
\draw[flow] (loc.north) |- (sum.east);
\draw[flow] (sum.north) -- ++(0,0.9) node[above,ann] {Output};
\node[smallann, anchor=west] at (2.78,3.35)
    {$O=\lambda\,\hat O_{\mathrm{hist}}+(1-\lambda)\,O_{\mathrm{local}}$};

\end{tikzpicture}